\documentclass[10pt,twocolumn,letterpaper]{article}

\usepackage[pagenumbers]{cvpr} 
\usepackage{graphicx}
\usepackage{amsmath}
\usepackage{amsbsy}
\usepackage{amssymb}
\usepackage{booktabs}
\usepackage{wrapfig}
\usepackage{tikz}
\usepackage{comment}
\usepackage{color}
\usepackage{graphicx}
\usepackage{multirow}
\usepackage{tabularx}
\usepackage{bbding}
\usepackage{bbm}
\usepackage{float}
\usepackage{xspace}
\usepackage[misc]{ifsym}
\usepackage{bbding}
\usepackage{pifont}
\usepackage{color,xcolor}
\usepackage{colortbl}
\usepackage{multicol}
\usepackage{stfloats}
\usepackage{subcaption}
\usepackage{textcomp}
\usepackage{tikz}
\newcommand*\circledM{%
  \tikz[baseline=(char.base)]{
    \node[shape=circle,draw,inner sep=0.16pt,minimum size=0.1em] (char) {\begin{scriptsize}$\mathcal{M}$\end{scriptsize}};
  }%
}

\usepackage{wasysym}
\newcommand{\cmark}{\ding{51}\xspace}%
\newcommand{\xmarkg}{\textcolor{lightgray}{\ding{55}}\xspace}%

\newcommand{\ours}{DsHmp\xspace}
\newcommand{\pub}[1]{\color{gray}{\scriptsize{[{#1}]}}}
\definecolor{cvprblue}{rgb}{0.21,0.49,0.74}
\usepackage[pagebackref,breaklinks,colorlinks,citecolor=cvprblue]{hyperref}

\def\paperID{2916} 
\def\confName{CVPR}
\def\confYear{2024}

\title{Decoupling Static and Hierarchical Motion Perception for \\Referring Video Segmentation}

\author{Shuting He$^1$
\qquad
Henghui Ding$^{1, 2}$~$^{\textrm{\Letter}}$\\
$^1$Nanyang Technological University
\qquad
$^2$Institute of Big Data, Fudan University
}

\begin{document}
\maketitle
\renewcommand{\thefootnote}{\fnsymbol{footnote}}
\footnotetext[0]{${\textrm{\Letter}}$ Corresponding author 
(henghui.ding@gmail.com).}

\begin{abstract}
Referring video segmentation relies on natural language expressions to identify and segment objects, often emphasizing motion clues. Previous works treat a sentence as a whole and directly perform identification at the video-level, mixing up static image-level cues with temporal motion cues. However, image-level features cannot well comprehend motion cues in sentences, and static cues are not crucial for temporal perception. In fact, static cues can sometimes interfere with temporal perception by overshadowing motion cues. In this work, we propose to decouple video-level referring expression understanding into static and motion perception, with a specific emphasis on enhancing temporal comprehension. Firstly, we introduce an expression-decoupling module to make static cues and motion cues perform their distinct role, alleviating the issue of sentence embeddings overlooking motion cues. Secondly, we propose a hierarchical motion perception module to capture temporal information effectively across varying timescales. Furthermore, we employ contrastive learning to distinguish the motions of visually similar objects. These contributions yield state-of-the-art performance across five datasets, including a remarkable \textbf{9.2\%} $\mathcal{J\&F}$ improvement on the challenging MeViS dataset. Code is available at \href{https://github.com/heshuting555/DsHmp}{https://github.com/heshuting555/DsHmp}.
\end{abstract}

\section{Introduction}

Referring video segmentation~\cite{seo2020urvos, MeViS, RefDAVIS,gavrilyuk2018actor} is a continually evolving area that lies at the crossroads of computer vision and natural language processing. This emerging realm of study is concentrated on segmenting and tracking specific objects of interest within video content, guided by natural language expressions. While it shares a historical connection with video object segmentation, it distinguishes itself by leveraging natural language expressions as guidance, which can offer motion-related cues like ``\textit{walking}'' and ``\textit{jumping}''. Notably, recent datasets like MeViS~\cite{MeViS} have emphasized the role of motion expressions in this evolving field, underscoring the significance of comprehending multi-modal motion information in this context.

\begin{figure}
    \centering
    \includegraphics[width=0.496\textwidth]{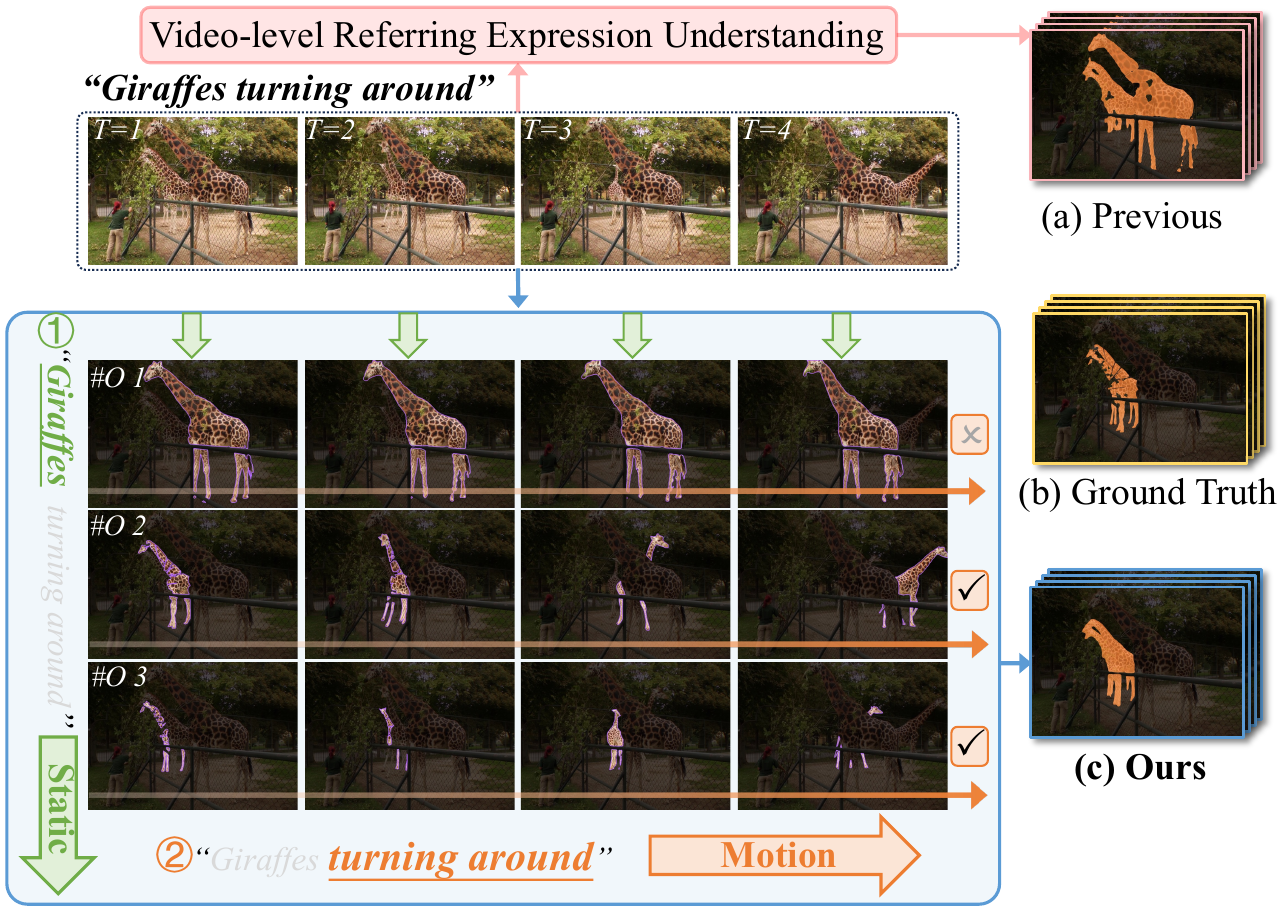} 
    \vspace{-7.36mm}
    \caption{Previous works treat a sentence as a whole and perform referring understanding at the video-level. However, image-level features struggle to understand motion cues, and static cues can sometimes disrupt temporal perception by overshadowing motion cues. We introduce a decoupling of static and motion perception, with a particular focus on enhancing temporal understanding.}
    \label{fig:teaser}
    \vspace{-2.6mm}
\end{figure}

Current approaches~\cite{referformer,MeViS,HTML,TempCD,OnlineRefer,SgMg,R2VOS} in referring video segmentation typically oversimplify the complex nature of language by reducing it to a single sentence embedding. For example, ReferFormer~\cite{referformer} and LMPM~\cite{MeViS} both employ the strategy of duplicating a single sentence embedding for multiple query embeddings within a Transformer architecture. This approach tends to overshadow the distinct importance and unique static or motion cues offered by the sentence.
For instance, consider the sentences: ``\textit{The little girl in red standing near the chair and drinking}'' and ``\textit{The little girl in red moving near the chair and drinking}''. These two sentences share 10 out of 11 words, leading to highly similar sentence embeddings, despite potentially referring to different targets. 
To address this issue, we propose to decouple image-level segmentation and temporal-level motion understanding. As shown in \cref{fig:teaser}, we let the given sentence focus on two distinct components: static and motion. The static cues are leveraged to identify potential candidates based on the static visual features present in each individual frame, where motion cues are not necessary. Then, the motion cues are used to pinpoint the target objects among the identified candidates by aligning them with temporal features observed throughout the video. In this way, static cues and motion cues perform their distinct and complementary roles, enhancing the comprehensive understanding of the referring expressions and videos.

A significant challenge in referring video segmentation is the precise capture and alignment of motions across the temporal domain. The motion cues provided by expressions may span a variable number of frames. For example, there are brief actions that occur over a few frames, like ``\textit{flying away}'', as well as long-term actions that persist throughout the entire video, such as ``\textit{walking from leftmost to rightmost}''. The unpredictability of the number of frames in which these actions occur greatly intensifies the challenge and complexity of capturing and comprehending motion. Recently, LMPM~\cite{MeViS} has introduced a method to capture motions using object tokens, offering computational efficiency and increasing the number of frames considered in temporal learning. 
However, LMPM tends to treat all frames uniformly and overlooks the distinctions between fleeting motions and long-term motions. In this work, we propose a hierarchical motion perception module to gradually comprehend temporal information based on object tokens, starting from short-term actions and progressing toward long-term actions. This module mimics the way humans understand videos by processing short clips and building an understanding of long-term concepts based on the recollection of short-term clips.

Furthermore, another challenge arises in differentiating between objects that exhibit nearly identical static appearances, such as two sheep that look very similar but have distinct motions. In such cases, the visual features extracted by the image encoder are highly alike, necessitating a significant reliance on the nuanced temporal differences to distinguish similar-looking objects. To address this challenge and enhance object discrimination using motion cues, we employ an object-wise contrastive learning. For a robust learning process, a memory bank is built to generate feature centroids for different objects, which greatly enhance the quality and quantity of positive and negative samples. We prioritize negative samples from the same category, like the three giraffes in \cref{fig:teaser}, as they are more likely to share a similar appearance and serve as challenging samples for distinguishing objects with distinct motions. This training objective effectively helps to emphasize the disparities in motion features for the similar-looking objects.

In summary, our main contributions are as follows:
\begin{itemize}
\setlength\itemsep{0.1em}
    \item We propose to decouple referring video segmentation into static perception and motion perception. Static perception focuses on grounding candidate objects on image-level based on static cues, while motion perception aims at understanding temporal context and identifying the target objects on temporal-level using motion cues.

    \item We propose a Hierarchical Motion Perception that effectively processes temporal motions, enabling the capture of motion patterns spanning various frame intervals.

    \item We leverage contrastive learning to acquire discriminative motion representations and enhance the model's ability to distinguish visually similar objects using motion cues.

    \item We achieve new state-of-the-art performance on five referring video segmentation datasets, with a particularly significant \textbf{9.2\% $\mathcal{J}\&\mathcal{F}$} improvement on the challenging MeViS dataset, showing the effectiveness of our method.

\end{itemize}

\section{Related Work}

\noindent\textbf{Referring Image Segmentation.}
Referring image segmentation aims to segment the target object within the image according to the given sentence~\cite{liu2017recurrent,li2018referring,feng2021encoder,ding2020phraseclick,margffoy2018dynamic,hu2016segmentation,MA3Net,ISFP}.
Its prevailing methods fall into two main categories: one-stage methods that perform end-to-end prediction and two-stage methods that involve instance segmentation followed by language-instance matching. For example, Hu et al.~\cite{hu2016segmentation} fuse visual and linguistic features and then conduct pixel-wise classification for mask prediction, representing a one-stage method. In contrast, Yu et al.~\cite{yu2018mattnet} use a instance segmentation model to detect all instances in the image and subsequently select the one that best aligns with the sentence, characterizing a two-stage approach.
More recently, the success of Transformer~\cite{vaswani2017attention,li2023transformer} has inspired a wave of research in referring image segmentation. Ding et al.~\cite{ding2021vision} first introduce Transformer into this domain and propose the Vision-Language Transformer (VLT). Subsequently, many Transformer-based methods have emerged~\cite{yang2021lavt, wang2022cris, kim2022restr,GRES,CGFormer,PolyFormer,yu2023zero,GroupRES,XDecoder,PartialRES,Hu_2023_ICCV,OpenVocaSurvey}.

\begin{figure*}[t]
    \centering
	\includegraphics[width=1.0\textwidth]{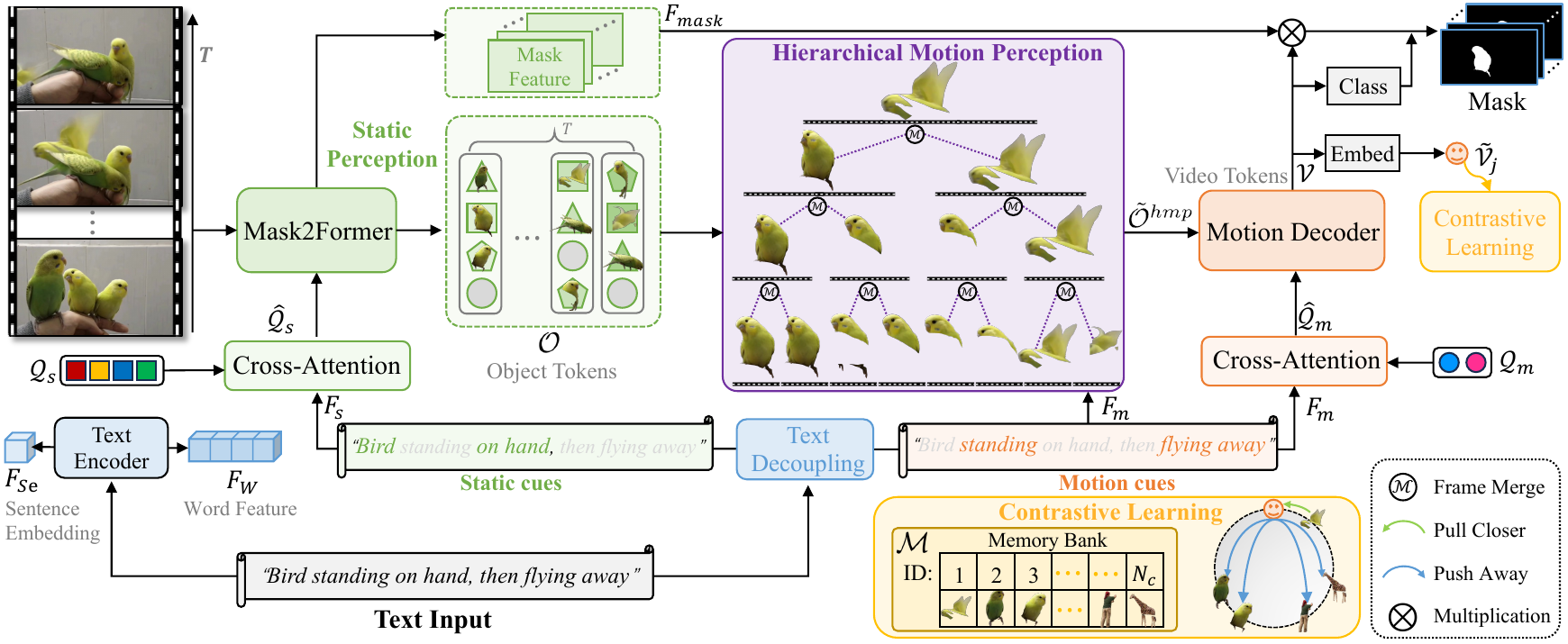}
	\vspace{-0.56cm}
    \caption{Overview of the proposed approach, named as DsHmp. We decouple the referring video segmentation to image-level static perception and temporal-level motion perception. We first employ Mask2Former to segment the possible objects according to static cues $F_s$. Then based on the object tokens $\mathcal{O}$ generated by Mask2Former, a hierarchical motion perception is employed to gradually comprehend temporal motions from short-term to long-term. Next, we employ a Motion Decoder to identify the target object according to motion cues $F_m$ and produce video tokens $\mathcal{V}$, which are used for mask predictions. Contrastive learning is applied on video tokens to help the model differentiate visually similar objects with distinct motion patterns.} 
	\vspace{-0.36cm}
	\label{fig:framework}
\end{figure*}

\noindent\textbf{Referring Video Segmentation.} Referring video segmentation aims to segment the target object within a given video according to a natural language expression~\cite{HTML, R2VOS, SgMg, TempCD,OnlineRefer,CMSA,CMPC,ding2021pminet,YOFO,ACAN,Clawcranenet,MANet,MLRL,visdrone}. 
This field is continually evolving with the introduction of A2D-Sentences~\cite{gavrilyuk2018actor}, Ref-DAVIS17~\cite{RefDAVIS}, Ref-YouTube-VOS~\cite{seo2020urvos}, and MeViS~\cite{MeViS}.
Many previous methods in referring video segmentation have primarily adapted referring image segmentation approaches to perform frame-by-frame target object segmentation, often overlooking the temporal dimension. For example, Khoreva \etal~\cite{RefDAVIS} ues the referring image segmentation method MAttNet~\cite{yu2018mattnet} for frame-level segmentation and then applied post-processing techniques to ensure temporal consistency. URVOS~\cite{seo2020urvos} and RefVOS~\cite{bellver2020refvos} utilize cross-modal attention for per-frame segmentation but do not leverage the temporal dimension. 
Despite their performance, these methods largely overlook the motion information inherent to videos. Some other works, like 
ReferFormer~\cite{referformer} and MTTR~\cite{MTTR}, employ the DETR-like structure in the RVOS field which simplifies the referring pipeline and achieves impressive performance.
More recently, based on complex video object segmentation dataset MOSE~\cite{MOSE}, MeViS~\cite{MeViS} dataset has been constructed to emphasize the significance of motion expressions and highlight the inadequacies of existing methods in comprehending the motion information present in languages and videos. In our work, the primary focus is on enhancing the understanding of motion cues in both visual and linguistic features.
\section{Approach}

The overview of the proposed approach, named DsHmp, is shown in \cref{fig:framework}. We first extract word feature $F_W$ and sentence embedding $F_{Se}$, and decouple the given sentence to static cues $F_s$ and motion cues $F_m$. With the static cues $F_s$ as queries, we employ a Mask2Former~\cite{mask2former} to extract object tokens $\mathcal{O}$ of potential candidate objects and mask feature $F_\text{mask}$ at image-level. Then the proposed Hierarchical Motion Perception (HMP) is conducted on the object tokens to progressively and hierarchically collect temporal information, generating motion-aware object tokens $\tilde{\mathcal{O}}^{hmp}$ with the guidance of motion cues $F_m$. Next, we employ the motion cues $F_m$ to identify the target object with a Motion Decoder and produce video tokens ${\mathcal{V}}$. Finally, the predicted masks are obtained by multiplying the video tokens $\mathcal{V}$ and mask features $F_\text{mask}$, and these with class scores higher than a threshold are selected as output. Contrastive learning is applied on the video tokens to enhance the model's ability to distinguish objects using motion cues. To facilitate contrastive learning, a memory bank $\mathcal{M}$ is established to store video tokens from various objects, ensuring a supply of high-quality negative samples.

\subsection{Decoupling Motion and Static Perception}

Existing approaches~\cite{referformer,MeViS,HTML,TempCD,OnlineRefer,SgMg,R2VOS} in referring video segmentation often oversimplify the complex nature of language by reducing it to a single sentence embedding. 
Meanwhile, commonly used visual backbones, such as Mask2Former~\cite{mask2former} and Video-Swin~\cite{videoswin}, primarily function as image-level or short-video-level (\eg, 5 frames) segmentation models.
They face a challenge in comprehending motion cues within a single sentence embedding. 

To address these challenges and effectively leverage the cues provided by expressions,  we propose to decouple the static perception and motion perception. Concerning language, we introduce a decoupling of the given expression into static and motion cues, which serve as cues for static perception and motion perception, respectively. In terms of visual processing, we employ Mask2Former~\cite{mask2former} to concentrate on extracting potential objects relevant to image-level static cues. Subsequently, the Hierarchical Motion Perception 
 (see \cref{HMP}) and Motion Decoder are responsible for capturing motion based on motion cues. This decoupling allows both sub-tasks, \ie, static perception and motion perception, to learn more comprehensively.

As shown in \cref{fig:framework}, given the sentence ``\textit{Bird standing on hand, then flying away},'' we employ an external tool~\cite{schuster2015generating} to identify nouns, adjectives, and prepositions in the sentence, yielding static cues such as ``\textit{bird, on, hand}.'' Meanwhile, we extract verbs and adverbs, obtaining motion cues like ``\textit{standing, flying away}.'' Consequently, we extract static words feature as $F_s \in \mathbb{R}^{K_s\times C}$ and motion words feature as $F_m \in \mathbb{R}^{K_m\times C}$, where $K_s$/$K_m$ represents the length of static/motion cues and $C$ is number of channels. It is worth noting that we add the sentence embedding $F_{Se}$ in both the motion and static features to provide a contextual understanding of the given expression.

\begin{figure*}[t]
    \centering
	\includegraphics[width=1.0\textwidth]{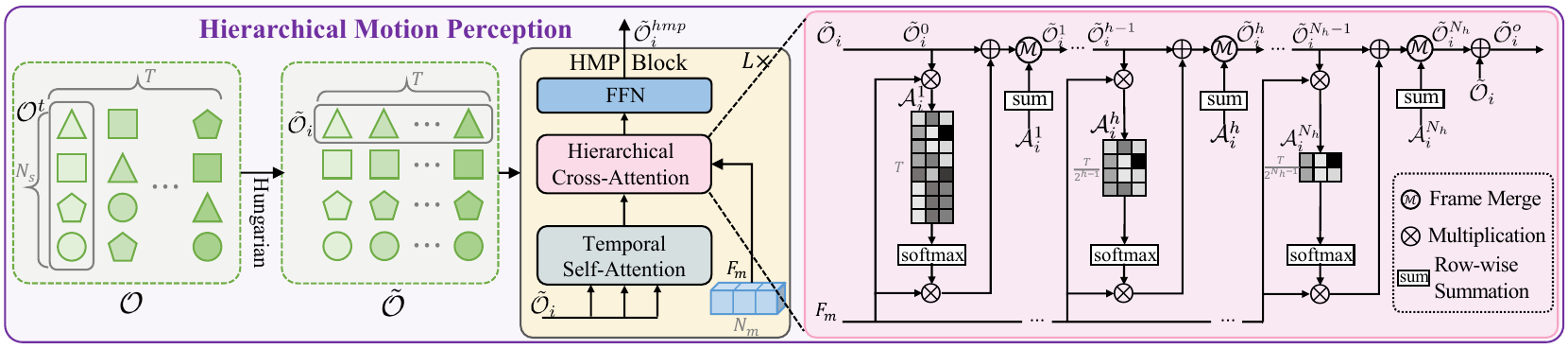}
	\vspace{-0.716cm}
    \caption{Architecture of the proposed Hierarchical Motion Perception (HMP). Hierarchical Motion Perception module effectively processes short-term and long-term motions, enabling the capture of motion patterns spanning various frame intervals.} 
	\vspace{-0.316cm}
	\label{fig:h_attention}
\end{figure*}

Furthermore, unlike the previous methods~\cite{MeViS,referformer}, which directly use language features as object query, we use cross-attention to inject static cues into the learnable static query:
\vspace{-2mm}\begin{equation}\label{eq:Qs}\vspace{-1mm}
   \hat{\mathcal{Q}}_s =  \mathcal{Q}_s+\mathrm{softmax}\left(\frac{\mathcal{Q}_s F_s^{T}}{\sqrt{C}}\right)F_s,
\end{equation}
where $\mathcal{Q}_s\in \mathbb{R}^{N_s\times C}$ is the $N_s$ initialized learnable static queries. After the cross-attention, $\hat{\mathcal{Q}_s}$ not only acquires a grasp of the objects data distributions but also captures specific static cues of the target. It enables the image-level segmentation of candidate objects via Mask2Former.

The motion cues $F_m$ are incorporated into the learnable motion query for Motion Decoder:
\vspace{-2mm}\begin{equation}\label{eq:Qm}\vspace{-2mm}
   \hat{\mathcal{Q}}_m =  \mathcal{Q}_m+\mathrm{softmax}\left(\frac{\mathcal{Q}_m F_m^{T}}{\sqrt{C}}\right)F_m,
\end{equation}
where $\mathcal{Q}_m\in \mathbb{R}^{N_m\times C}$ is the $N_m$ initialized learnable motion queries. Following the cross-attention, $\hat{\mathcal{Q}}_m$ gains specific motion cues related to the target. This facilitates the identification of target objects through the Motion Decoder. In this way, static cues and motion cues play distinct and complementary roles, enhancing the comprehensive understanding of the referring expressions and videos.

\subsection{Hierarchical Motion Perception}\label{HMP}

A significant challenge in referring video segmentation is precisely capturing and aligning motions across different timeframes. Expressions can provide motion cues that vary in the number of frames they span, \eg, fleeting or long-term motions. 
Inspired by LMPM~\cite{MeViS} that captures motions via object tokens, we propose a Hierarchical Motion Perception (HMP) module based on object tokens. This module progressively analyzes short-term and long-term motions, mirroring human video comprehension by forming an understanding of long-term concepts through the recollection of short-term clips.

\cref{fig:h_attention} illustrates the architecture of the proposed HMP module. It takes in the object tokens $\{\mathcal{O}^{t} | t\in~\![1, T], \mathcal{O}^{t} \in\mathbb{R}^{N_s\times~\!C}\}$, which are the $N_s$ candidate objects' tokens generated by the Mask2Former for each of the $T$ frames, and it outputs the motion-aware tokens 
$\tilde{\mathcal{O}}^{hmp}$. In the HMP module, the Hungarian matching algorithm \cite{hungarian} is first employed to match $\mathcal{O}$ of adjacent frames, as is done in \cite{minvis}:
\vspace{-1mm}\begin{equation}\vspace{-1mm}
  \left\{
  \begin{aligned}
  \tilde{\mathcal{O}}^{t} & = \text{Hungarian}(\tilde{\mathcal{O}}^{t-1},\mathcal{O}^{t}), \quad t\in[2,T] \\
  \tilde{\mathcal{O}}^{t} & = \mathcal{O}^{t}, \quad t=1
  \end{aligned}
  \right.
\label{eq:1},
\end{equation}
where $\tilde{\mathcal{O}}$ is the matched object tokens and can be considered as the tracking result with noise. As such, we can obtain $N_s$ object trajectories, \ie, $\{\tilde{\mathcal{O}}_{i} | i \in [1, N_s], \tilde{\mathcal{O}}_{i} \in  \mathbb{R}^{ T\times C } \}$, where $\tilde{\mathcal{O}}_{i}$ is a single object trajectory.
As shown in the middle part of \cref{fig:h_attention}, the proposed Hierarchical Motion Perception (HMP) module is composed of $L$ HMP blocks that are cascaded together. Each block consists of three main components: temporal self-attention, hierarchical cross-attention, and FFN layer. Temporal self-attention is utilized to grasp long-term motion, while hierarchical cross-attention is used to progressively and hierarchically gather temporal information from the short-term to the long-term.
To perform the hierarchical cross-attention, we first highlight frames containing the target motions by calculating the similarity between the motion feature $F_m$ and each object trajectory $\tilde{\mathcal{O}}_i^{h-1}$ from last hierarchical stage using the following equation:
\vspace{-1mm}\begin{equation}\label{eq:a1}\vspace{-1mm}
\small
\mathcal{A}_i^h = \mathrm{softmax}\left(\frac{\tilde{\mathcal{O}}_i^{h-1} {F_m}^{T}}{\sqrt{C}}\right),
\end{equation}
where $h\in [{1},N_h]$ denotes the stage number of hierarchical operation and $\tilde{\mathcal{O}}_i^0 = \tilde{\mathcal{O}}_i$. 
$\mathcal{A}_i^h \in \mathbb{R}^{ T_h\times K_m }$ represents the attention map for the $T_h$ frames of object trajectory $\tilde{\mathcal{O}}_i^{h-1}$ and the $K_m$ motion cues, where $T_h=\frac{T}{2^{h-1}}$. 
The softmax operation is performed on the $T_h$ axis.
Next, we obtain the motion feature-enriched object feature $\hat{\tilde{\mathcal{O}}}_{i}^h$ by incorporating the related motion cues, \ie,
\vspace{-1mm}\begin{equation}\label{eq:h1}\vspace{-1mm}
\small
\hat{\tilde{\mathcal{O}}}_{i}^h = \tilde{\mathcal{O}}_i^{h-1} + \frac{\mathcal{A}_i^h}{\hat{\mathcal{A}}_i^h} F_m,
\end{equation}
where $\hat{\mathcal{A}}_i^h= \sum_k\mathcal{A}_i^{h,(t,k)} \in \mathbb{R}^{T_h}$ can be regarded as frame weight by summing the effect of $K_m$ motion cues on each of the $T_h$ frames. This weight signifies the importance of each individual frame's token within the trajectory spanning over $T_h$ frames concerning the motion cues.

Subsequently, we engage in token merging to accumulate short-term motion information and create merged tokens for higher-level understanding within the hierarchical framework. 
With the motion feature-enriched object feature $\hat{\tilde{\mathcal{O}}}_{i}^h$ and their corresponding importance weight $\hat{\mathcal{A}}_i^h$, we apply a frame merging operation to combine adjacent two tokens into one token using the following equation:
\begin{equation}\label{eq:merge}
\small
   {\tilde{\mathcal{O}}_i^{h}} =  
   \begin{tiny}\circledM\end{tiny}(\hat{\tilde{\mathcal{O}}}_{i}^h,
\hat{\mathcal{A}}_i^h),
\end{equation}
where $\tilde{\mathcal{O}}^{h}_i  \in \mathbb{R}^{ \frac{T_h}{2}\times C }$. 
\begin{tiny}\circledM\end{tiny}~is a token merging operation that blends two tokens of neighboring frames into a single one using a weighted average based on $\hat{\mathcal{A}}_i^h$. 
The merging operation reduces redundant tokens and enhances those associated with motion cues. The merged trajectory ${\tilde{\mathcal{O}}_i^{h}}$ is used as input for the next stage to generate ${\tilde{\mathcal{O}}_i^{h+1}} \in \mathbb{R}^{ \frac{T_h}{4}\times C }$ with a larger temporal context view.

The operations from \cref{eq:a1} to \cref{eq:merge} are iteratively performed for a total of $N_h$ times, as shown in \cref{fig:h_attention}. This gradual merging of tokens and expansion of the temporal scope forms a hierarchical motion perception, transitioning from short-term to long-term motion understanding. 
The output of the hierarchical cross-attention is ${\tilde{\mathcal{O}}_i^{o}}$, and it is obtained by:
\begin{equation}\label{eq:add}
\small
   {\tilde{\mathcal{O}}_i^{o}} =  
{\tilde{\mathcal{O}}_i}+{\tilde{\mathcal{O}}_i^{N_h}},
\end{equation}
where ${\tilde{\mathcal{O}}_i^{N_h}} \in \mathbb{R}^{ \frac{T}{2^{N_h}}\times C}$ and is expanded to match the dimensions of $\tilde{\mathcal{O}}_i$ for summation.
$\tilde{\mathcal{O}}_i^{o} \in \mathbb{R}^{ T\times C}$ is then fed to the FFN to generate this block's output,  which serves as the input for the next HMP block.

The final output of the Hierarchical Motion Perception module, denoted as $\tilde{\mathcal{O}}^{hmp}$, is used as the key and value inputs to the Motion Decoder for the identification of target objects, along with the query $\hat{\mathcal{Q}}_m$ generated by \cref{eq:Qm}. 
The Motion Decoder produces video tokens for target objects, denoted as ${\mathcal{V}}\in \mathbb{R}^{N_m\times C}$. With the motion-aware object tokens $\tilde{\mathcal{O}}^{hmp}$ generated by Hierarchical Motion Perception, Motion Decoder is able to more effectively understand the motion information conveyed by the language.

\subsection{Contrastive Learning}

Although the proposed hierarchical motion perception simplifies the motion identification for the Motion Decoder, the presence of objects with highly similar appearances can still pose challenges and lead to confusion in the identification process.
To address this challenge, we apply contrastive learning on the output of Motion Decoder, video token $\mathcal{V}$. This approach enhances the model's capacity to differentiate similar-looking objects via motion cues.

\noindent$\bullet$~\textbf{Vanilla Samples Selection.} In contrastive learning, the choice of positive and negative samples is crucial~\cite{he2020momentum, wu2018unsupervised}. We select the video token\footnote{For MeViS~\cite{MeViS} dataset with multiple target objects, we average the matched video tokens to represent their collective characteristics.} with the lowest cost to the ground truth {as the anchor}, while video tokens of other objects in the mini-batch serve as negative samples. However, this straightforward approach faces two issues: 1) lack of corresponding positive samples, and 2) insufficient negative samples {limited by mini-batch size}, which significantly impact the final outcome~\cite{kalantidis2020hard}. To address these issues, we introduce a memory bank to store more video tokens.

\noindent$\bullet$~\textbf{Memory Bank.} Since there are vast numbers of video tokens in our training process, directly storing all the video tokens, like a traditional memory bank~\cite{chen2020simple}, severely slows down the learning process. Therefore, we choose to maintain a video token centroid for per target. We introduce a memory bank $\mathcal{M} \in \mathbb{R}^{N_c\times C}$ to gather representative video token centroid for each target, with each element denoting the feature centroid of projected video token $\in \mathbb{R}^C$ from the contrastive head. $N_c$ is the number of target objects in dataset. 
Specifically, given a video, we utilize the projected feature of anchor video token $\tilde{\mathcal{V}}_j$ to update the corresponding target object centroid feature of $\mathcal{M}$:
\vspace{-0.36mm}\begin{equation}\vspace{-0.36mm}
    \mathcal{M}_{[\tilde{\mathcal{V}}_j]} = \beta \mathcal{M}_{[\tilde{\mathcal{V}}_j]} + (1-\beta) \tilde{\mathcal{V}}_j,
\end{equation}
where $[\tilde{\mathcal{V}}_j]$ is an index mapping from $\tilde{\mathcal{V}}_j$ to its corresponding target index in the memory bank. The hyperparameter~$\beta$ controls the update speed. This way brings two advantages: 1) Storing the centroid feature of target objects instead of each target object feature greatly saves memory consumption. 2) The centroid feature is more representative and robust to encompass the static and motion cues, contributing to the construction of a well-structured feature space.

We apply an object-wise contrastive learning as follows:
\begin{equation}\label{eq:con}
\small
    \mathcal{L}_{con} = -\log\frac{\mathrm{exp}({\tilde{\mathcal{V}}_j}  \cdot {m^+}/\tau)}{\mathrm{exp}({\tilde{\mathcal{V}}_j}\cdot{m^+}/\tau)+\sum_{m^-\in \mathbf{N}}\mathrm{exp}({\tilde{\mathcal{V}}_j}\cdot{m^-}/\tau)},
\end{equation}
where $\tau$ is a temperature hyperparameter. Positive sample $m^+$ is the feature centroid of the target object that $\tilde{\mathcal{V}}_j$ belongs to, \ie $\mathcal{M}_{[\tilde{\mathcal{V}}_j]}$.
$\mathbf{N}$ is the collection of $N_n$ negative samples which come from different objects in $\mathcal{M}$. We prioritize negatives belonging to the same category in the same video, as they are more likely to have a similar appearance and serve as the challenging samples in contrastive learning of distinct motions. $\mathcal{L}_{con}$ is computed after $N_i$ iterations to ensure a stable training process.

\subsection{Training Objective}
Following~\cite{MeViS,VITA}, we employ the match loss $\mathcal{L}_{f}$ between per-frame outputs and frame-wise ground truth, along with $\mathcal{L}{v}$ as the video-level loss with video-level ground truth.
The total training objective to optimize the model is:
$\mathcal{L}_{train} = \mathcal{L}_{{f}} + \mathcal{L}_{{v}} +\lambda_{con} \mathcal{L}_{{con}}$,
where $\lambda_{con}$ is used for balancing the contrastive loss $\mathcal{L}_{{con}}$.
\section{Experiments}

\subsection{Datasets and Evaluation Metrics}
\noindent\textbf{Dataset.}
The proposed approach, named as \textbf{\ours}, is evaluated on five video datasets: MeViS~\cite{MeViS}, Ref-YouTube-VOS~\cite{seo2020urvos}, Ref-DAVIS17~\cite{RefDAVIS}, A2D-Sentences~\cite{gavrilyuk2018actor}, and JHMDB-Sentences~\cite{JHMDB}. MeViS is a newly established dataset that is targeted at motion information analysis and consists of 2,006  videos and 28K annotations. The Ref-YouTube-VOS stands out as the most extensive R-VOS dataset, comprising 3,978 videos and approximately 13K annotations.
Ref-DAVIS17 builds upon DAVIS17~\cite{davis2017}, enriched with linguistic annotations for diverse objects, and offers 90 videos.
A2D-Sentences, designed for actor and action segmentation, encompasses over 3.7K videos paired with 6.6K action annotations. Meanwhile, JHMDB-Sentences provides 928 videos, each with a description, spread across 21 unique action categories.

\noindent\textbf{Evaluation Metrics.}
Unless otherwise specific, the evaluation metrics we used are: region similarity $\mathcal{J}$ (average IoU), contour accuracy $\mathcal{F}$ (mean boundary similarity), and their average 
\( \mathcal{J} \)\&\( \mathcal{F} \).
The evaluations are conducted using the official code or online platforms.
For A2D-Sentences and JHMDB-Sentences, we employ mAP, overall IoU (oIoU), and mean IoU (mIoU) as the evaluation metrics.

\subsection{Implementation Details}
For experiments on MeViS dataset, we follow the default setting of \cite{MeViS}. Specifically, we train the models directly on MeViS without any pre-training on RefCOCO/+/g~\cite{RefCOCO2,RefCOCO}. The training spans 50,000 iterations using the AdamW optimizer~\cite{loshchilov2017adamw} with a learning rate set at 0.00005.
For experiments on YouTube-VOS/A2D-Sentences, following \cite{MTTR,referformer}, the experiments begin with pre-training on RefCOCO/+/g~\cite{RefCOCO2,RefCOCO} and then undergo main training. Besides, models trained on the Ref-YouTube-VOS/A2D-Sentences training set are evaluated directly on the val set of Ref-DAVIS17/JHMDB-Sentences without the use of additional post-processing techniques. 
During the pre-training phase, we train the model with 300,000 iterations. In the main training phase, we train with 50,000 iterations.
All experiments use RoBERTa~\cite{liu2019roberta} as the text encoder.
All frames are cropped to have the longest side of 640 pixels and the shortest side of 360 pixels during training and evaluation.
For hyperparameters, we set values for $N_s$, $N_m$, $N_h$, $N_n$, $N_i$, $\beta$, $\tau$, $\lambda_{con}$ at 20, 10, 3, 100, 10,000, 0.2, 0.07, 0.5, respectively. 

\subsection{Ablation Study}

\begin{table}[t!]
\centering
\footnotesize
\setlength{\tabcolsep}{9.36pt}
\begin{tabular}{c|c c c| c c c }
\rowcolor[gray]{.9}
\hline 
&\multicolumn{3}{c|}{Components} & \multicolumn{3}{c}{Results} \\
\rowcolor[gray]{.9}

Index&DS & HMP & CL & \( \mathcal{J} \)\&\( \mathcal{F} \) &  \( \mathcal{J} \) & \( \mathcal{F} \)  \\
\hline 
\hline
0& \xmarkg& \xmarkg & \xmarkg & 39.7 & 36.6 &  42.8 \\ 
1&\cmark  &\xmarkg  & \xmarkg & 42.5 & 39.4 & 45.6  \\
2& \xmarkg & \cmark &\xmarkg  & 43.8 & 40.7 & 46.9  \\
3&\xmarkg &\xmarkg & \cmark & 42.1 &  39.0& 45.2 \\ 
4&\cmark & \cmark &\xmarkg  &45.1 &  41.8 &48.4   \\
5&\cmark & \xmarkg &\cmark  &43.9  &40.8  &47.0   \\
6&\xmarkg & \cmark &\cmark  &44.9  & 41.7 & 48.1  \\
\rowcolor{cyan!10}7&\cmark & \cmark & \cmark & \textbf{46.4} &  \textbf{43.0}  &\textbf{49.8}\\
\hline 
\end{tabular}
\vspace{-3mm}
\caption{Ablation study of our method on MeViS dataset.
DS, HMP, and CL denote components of decoupling sentence, hierarchical motion perception, and contrastive learning, respectively.} \label{tab:abmodule} 
\vspace{-3mm}
\end{table}

 \begin{table*}[t!]
 \footnotesize
    \begin{center}
    \setlength{\tabcolsep}{3.96pt}
    \begin{subtable}[t]{0.28\linewidth}
        \centering
        \begin{tabular}{c|ccc}
         \rowcolor[gray]{.9}
        \hline
        Input Query &$\mathcal{J}$\&$\mathcal{F}$& $\mathcal{J}$ & $\mathcal{F}$\\
        \hline
        \hline
        $F_{Se}$ &  44.9& 41.7 &48.1    \\
        DS w/o $F_{Se}$ & 45.6 & 42.1 & 49.1 \\
        DS w/o $\mathcal{Q}_s$/$\mathcal{Q}_m$ & 45.9 & 42.3 &49.5 \\
 	  \rowcolor{cyan!10}DS &  \textbf{46.4} &  \textbf{43.0}  &\textbf{49.8}  \\

        \hline
        \end{tabular}
        \vspace{-1mm}
        \caption{Different input query variations. }
        \label{table:3a}
    \end{subtable}
\hfill
    \setlength{\tabcolsep}{3.96pt}
    \begin{subtable}[t]{0.23\linewidth}
        \centering
        \begin{tabular}{c|ccc}
         \rowcolor[gray]{.9}
        \hline
        $N_h$ &$\mathcal{J}$\&$\mathcal{F}$& $\mathcal{J}$ & $\mathcal{F}$\\
        \hline
        \hline
        0 & 43.9& 40.8 & 47.0    \\
 	  1 &  45.0& 41.8 &48.2  \\
        2 & 45.8 & 42.3 & 49.3 \\
        \rowcolor{cyan!10}3 &\textbf{46.4} &  \textbf{43.0}  &\textbf{49.8}  \\
        \hline
        \end{tabular}
        \vspace{-1mm}
        \caption{Different hierarchical stages. }
        \label{table:3b}
    \end{subtable}
\hfill
    \setlength{\tabcolsep}{3.96pt}
    \begin{subtable}[t]{0.2\linewidth}
        \centering
        \begin{tabular}{c|ccc}
         \rowcolor[gray]{.9}
        \hline
         $N_n$ &$\mathcal{J}$\&$\mathcal{F}$& $\mathcal{J}$ & $\mathcal{F}$ \\
        \hline
        \hline
 	  0 &  45.1 &  41.8 &48.4 \\
        10 & 45.4 &  41.9 & 48.9\\
        \rowcolor{cyan!10}100 & \textbf{46.4} &  \textbf{43.0}  &\textbf{49.8}  \\
        200 & 46.5 & 43.1 & 49.8 \\
        \hline
        \end{tabular}
        \vspace{-1mm}
        \caption{Different negative samples. }
        \label{table:3c}
    \end{subtable}
\hfill
    \setlength{\tabcolsep}{3.96pt}
    \begin{subtable}[t]{0.2\linewidth}
        \centering
        \begin{tabular}{c|ccc}
         \rowcolor[gray]{.9}
        \hline
        Tokens &$\mathcal{J}$\&$\mathcal{F}$& $\mathcal{J}$ & $\mathcal{F}$\\
        \hline
        \hline
       $\mathcal{O}$  &  43.8 & 40.7 &46.9    \\
 	  \rowcolor{cyan!10}$\tilde{\mathcal{O}}$ &  \textbf{46.4} &  \textbf{43.0}  &\textbf{49.8}  \\
        \hline
        \end{tabular}
        \vspace{-1mm}
        \caption{Effect of Hungarian match. }
    \end{subtable}
\end{center}
\vspace{-6mm}
  \caption{
    \textbf{Ablation studies of different architecture designs} on MeViS. 
    }
    \vspace{-3mm}
     \label{tab:hyper_parameter}
\end{table*}

Since the main focus of this paper is exploiting motion information, we conduct ablation study on MeViS~\cite{MeViS}.

\noindent\textbf{Module Effectiveness}.
We conduct ablation experiments to evaluate the effectiveness of different components.
As shown in~\cref{tab:abmodule}, the inclusion of Decoupling Sentence (DS, index 1) leads to a performance improvement of \textbf{2.8\%} $\mathcal{J}$\&$\mathcal{F}$ compared to vanilla baseline (index 0), which is adopted from LMPM~\cite{MeViS} with our reproduction. The introduction of DS enhances the model's ability to comprehensively learn static and motion cues for both image-level and temporal-level understanding.
Then, HMP (index 2) is used to capture motion information at multiple temporal granularities, encompassing both short-term and long-term motions. HMP improves the performance by \textbf{4.1\%} $\mathcal{J}$\&$\mathcal{F}$, highlighting the significance of motion understanding for referring video segmentation.
Next, we present Contrastive Learning (CL, index 3) to construct discriminative motion representations enhancing the model's ability to distinguish similar-looking objects. Utilizing CL improves the $\mathcal{J}$\&$\mathcal{F}$ by \textbf{2.4\%}. When integrating all the components together (index 7), referred to as \ours, we observe a substantial improvement and achieve new state-of-the-art performance of 46.4\% $\mathcal{J}$\&$\mathcal{F}$ on the challenging MeViS dataset, demonstrating the effectiveness of the proposed method.

\noindent\textbf{Importance of sentence decoupling}.~In \cref{tab:hyper_parameter} (a), we study the impact of sentence decoupling on the input queries to MaskFormer and Motion Decoder. Utilizing only the basic sentence embedding $F_{Se}$ like ReferFormer~\cite{referformer} results in a 1.5\% $\mathcal{J}\&\mathcal{F}$ decrease, 
demonstrating that relying solely on sentence embedding limits discriminative capability and risks overlooking key cues.
Meanwhile, solely using the sentence decoupling query without the sentence embedding $F_{Se}$ leads to a 0.8\% $\mathcal{J}$\&$\mathcal{F}$ decrease, which is due to the lack of sentence context. 
DS w/o $\mathcal{Q}_s$/$\mathcal{Q}_m$ uses $F_s$ and $F_m$ directly, bypassing the integration of static or motion cues into the learnable query, leading to a 0.5\% drop in $\mathcal{J}\&\mathcal{F}$. This underscores the importance of incorporating a learnable query to grasp the global dataset distribution.
These findings show that while sentence embeddings are vital for maintaining language comprehension, relying solely on them does not facilitate a thorough learning process. Besides, decoupling sentence into static and motion cues is helpful for enhancing referring video segmentation.

\noindent\textbf{Number of hierarchical stages $N_h$ in HMP}.~\cref{tab:hyper_parameter} (b) shows results obtained with varying numbers of hierarchical stages. For $N_h$ = 0, only vanilla temporal self-attention is applied. For $N_h$ = 1, the hierarchical cross-attention mechanism is introduced to capture motion cues at per-frame level. With increasing $N_h$, the attention mechanism is refined to capture a larger temporal context. This hierarchical merging of tokens and extension of the temporal scope facilitates a transition from short-term to long-term motion understanding. To effectively capture motion information across multiple levels of granularity, we have configured the hierarchical stages at 3 to achieve the best result.

\noindent\textbf{Hungarian match in HMP}.~\cref{tab:hyper_parameter} (d) shows the necessity of adding Hungarian in HMP. Without Hungarian match, the calculation of motion information and token merge may not align for the same object, adversely impacting performance and resulting in a 2.6\% drop in $\mathcal{J}\&\mathcal{F}$.

\noindent\textbf{Number of negative samples $N_n$}.~In \cref{tab:hyper_parameter} (c), we report results with different numbers of negative samples used in contrastive loss in \cref{eq:con}. When only using 10 negative samples, which can be achieved within a mini-batch without memory bank, the performance is improved by 0.3\% $\mathcal{J}\&\mathcal{F}$ only. Increasing $N_n$ to 100 brings 1.3\% $\mathcal{J}\&\mathcal{F}$ performance gain, demonstrating the necessity of using a memory bank to store more samples. Employing more negative samples in \cref{eq:con} contributes to the establishment of a discriminative and comprehensive motion representation space yet consumes more computing resources. Striking a balance between accuracy and efficiency, we set $N_n$ to 100.

\begin{figure}[t]
    \centering
	\includegraphics[width=0.48\textwidth]{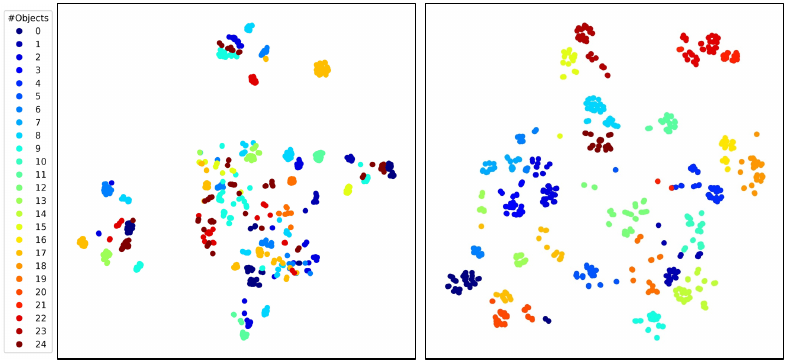}
	\vspace{-0.66cm}
    \caption{Visualization of features learned w/o CL (left) and w/ CL (right). Features are colored according to class labels. As seen, the proposed CL brings a well-structured video token feature space.} 
	\vspace{-0.16cm}
	\label{fig:tsne}
\end{figure}

\noindent\textbf{t-SNE visualization}. In \cref{fig:tsne}, we employ t-SNE~\cite{tsne} to visualize video token distribution with and without our contrastive learning across 25 different objects, each described by multiple language expressions. Without contrastive learning, tokens for the same object diverge due to language diversity, leading to overlap between tokens of similar-looking objects. Contrastive learning, on the other hand, brings tokens of the same target object closer and separates them from other objects, enhancing the model's ability to distinguish motions of visually similar objects.

\begin{table}[t]
   \centering
   \footnotesize
   \setlength{\tabcolsep}{3.mm}
{\begin{tabular}{r|c|ccc}
 \rowcolor[gray]{.9}
 \hline
         Methods& Reference&$\mathcal{J}$\&$\mathcal{F}$  & $\mathcal{J}$ & $\mathcal{F}$ \\
         \hline
           \hline
         URVOS~\cite{seo2020urvos}&\pub{ECCV'20} &27.8&25.7&29.9\\
         LBDT~\cite{LBDT}& \pub{CVPR'22}&29.3&27.8&30.8\\
         MTTR~\cite{MTTR}&\pub{CVPR'22} &30.0&28.8&31.2\\
         ReferFormer~\cite{referformer} &\pub{CVPR'22} &31.0&29.8&32.2\\
         VLT+TC~\cite{vltpami}& \pub{TPAMI'22}&35.5&33.6&37.3\\
         LMPM~\cite{MeViS} &\pub{ICCV'23}  &{37.2} &{34.2}& {40.2}\\       
         \rowcolor{cyan!10}\textbf{\ours} (ours)& \pub{CVPR'24} &\textbf{46.4} &  \textbf{43.0}  &\textbf{49.8}\\
         \specialrule{.1em}{.05em}{.05em}
      \end{tabular}}%
      \vspace{-3mm}
       \caption{Comparison on MeViS.}
   \label{tab:MeViS}%
   \vspace{-3.6mm}
\end{table}%

\subsection{Comparison with State-of-the-Art Methods}

\noindent\textbf{MeViS}~\cite{MeViS}. In \cref{tab:MeViS}, we evaluate the proposed approach {\ours} on the newly released motion expression video segmentation dataset MeViS. Following~\cite{MeViS}, we use  Swin-Tiny as the backbone. \ours achieves superior performance compared to other state-of-the-art methods and surpasses the previous state-of-the-art LMPM~\cite{MeViS} by a remarkable \textbf{9.2\%} $\mathcal{J}$\&$\mathcal{F}$. These results demonstrate the effectiveness of our approach \ours in capturing motion information.

\noindent\textbf{Ref-YouTube-VOS}~\cite{seo2020urvos} \& \textbf{Ref-DAVIS17}~\cite{RefDAVIS}. In \cref{tab:ytvos_davis},
we report results on Ref-YouTube-VOS and Ref-DAVIS17. Our approach exceeds existing methods on the two datasets across all metrics.
On Ref-YouTube-VOS, \ours with the Video-Swin-Tiny backbone achieves 63.6\% \( \mathcal{J}\)\&\( \mathcal{F}\), which is {1.2\%} higher than the previous state-of-the-art SOC~\cite{SOC}. When a larger backbone is used, \ie, Video-Swin-Base, the performance of \ours further improves to 67.1\% \( \mathcal{J} \)\&\( \mathcal{F} \), consistently outperforming all other methods by more than {1.1\%}.
On Ref-DAVIS17, our approach achieves 64.9\% \( \mathcal{J} \)\&\( \mathcal{F} \) and surpasses SOC~\cite{SOC} by {0.7\%}.
The performance gains on these two datasets are relatively modest compared to MeViS, mainly because the datasets may include sentences with image-level descriptions for the first frame and not strictly require motion expressions. Despite this, our method maintains state-of-the-art performance, underscoring its generalizability and effectiveness.

\noindent\textbf{A2D-Sentences \& JHMDB-Sentences}~\cite{gavrilyuk2018actor}. 
We further evaluate the proposed approach \ours on A2D-Sentences and JHMDB-Sentences in \cref{tab:A2D_JHMDB}. Following~\cite{referformer}, the models are first pre-trained on RefCOCO/+/g and then fine-tuned on A2D-Sentences. JHMDB-Sentences is used for evaluation only. 
The proposed \ours achieves new state-of-the-art performance and outperforms the nearest competitor SgMg~\cite{SgMg} by {1.3\%} and {0.8\%} mAP on A2D-Sentences and JHMDB-Sentences, respectively.

\begin{figure*}[t]
    \centering
	\includegraphics[width=1.0\textwidth]{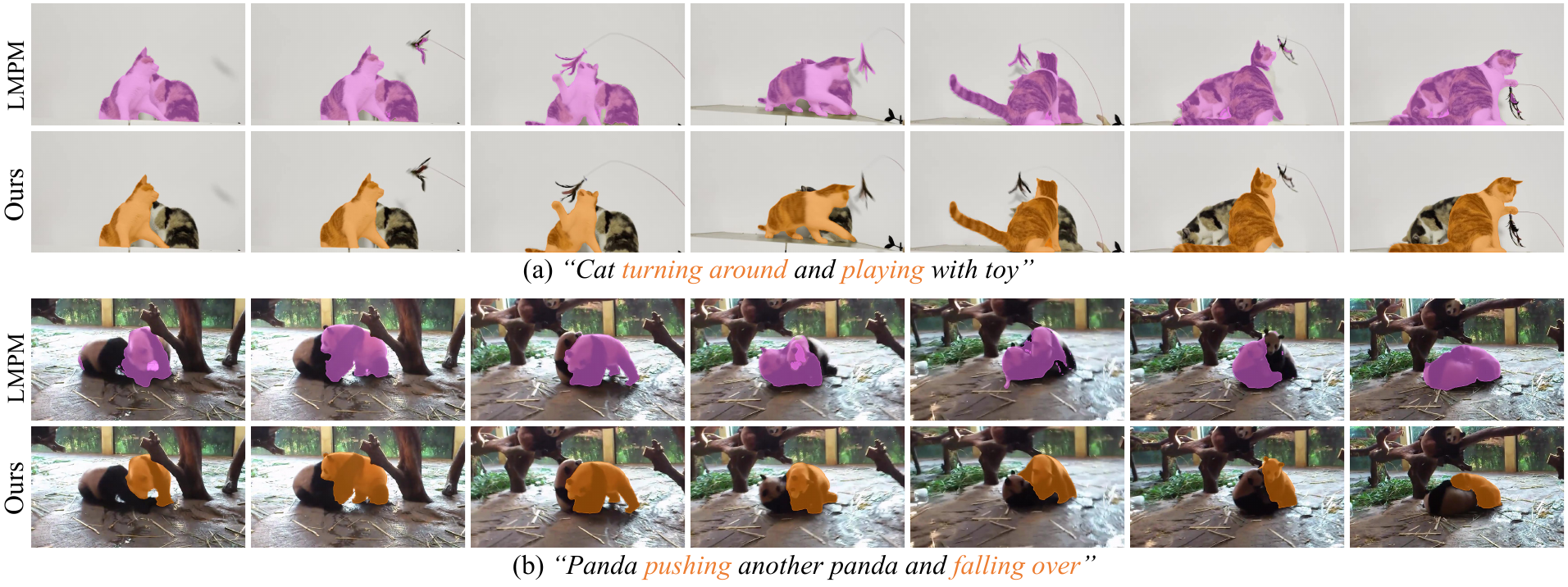}
 \vspace{-5.6mm}
    \caption{Visualization results of complex and motion language descriptions on MeViS. \textcolor{orange}{Orange} masks represent positive segmentation results and \textcolor{magenta}{pink} masks denote the negatives. Our \ours can capture temporal information effectively across varying timescales.} 
	\label{fig:visualization}
 \vspace{-3mm}
\end{figure*}

\begin{table}[t!]
\setlength{\tabcolsep}{2.5pt}
\centering
\footnotesize
\begin{tabular}{r | c |c c c | c c c}
 \rowcolor[gray]{.9}
\hline
&  &  \multicolumn{3}{c |}{Ref-YouTube-VOS} & \multicolumn{3}{c}{Ref-DAVIS17} \\
\rowcolor[gray]{.9} 
Method &Reference & \( \mathcal{J} \)\&\( \mathcal{F} \) & \( \mathcal{J} \) & \( \mathcal{F} \)  &  \( \mathcal{J} \)\&\( \mathcal{F} \) & \( \mathcal{J} \) & \( \mathcal{F} \) \\
\hline 
\hline
\multicolumn{8}{c}{Video-Swin-Tiny}\\
\hline

ReferFormer~\cite{referformer} & \pub{CVPR'22} & 59.4 & 58.0 & 60.9  & 59.6 & 56.5 & 62.7  \\
HTML~\cite{HTML} &\pub{ICCV'23} & 61.2 &59.5 &63.0&-&-&-\\
R$^2$-VOS~\cite{R2VOS}& \pub{ICCV'23} & 61.3 &59.6 &63.1 &-&-&-\\
SgMg~\cite{SgMg}  & \pub{ICCV'23} & {62.0} & {60.4} & {63.5}  & {61.9} & {59.0} & {64.8} \\ 
TempCD~\cite{TempCD}&\pub{ICCV'23} &62.3&60.5& 64.0& 62.2 &59.3& 65.0 \\
SOC~\cite{SOC} &\pub{NIPS'23}&62.4& 61.1& 63.7 &63.5 &60.2 &66.7\\
\rowcolor{cyan!10}\textbf{\ours} (ours)&\pub{CVPR'24}&\textbf{63.6} &\textbf{61.8} & \textbf{65.4} & \textbf{64.0}& \textbf{60.8}&\textbf{67.2} \\
\hline
\multicolumn{8}{c}{Video-Swin-Base}\\
\hline
ReferFormer~\cite{referformer} & \pub{CVPR'22} & 62.9 & 61.3 & 64.6 & 61.1 & 58.1 & 64.1 \\
OnlineRefer~\cite{OnlineRefer} & \pub{ICCV'23} &62.9& 61.0 &64.7 &62.4 &59.1 &65.6\\
HTML~\cite{HTML} &\pub{ICCV'23} & 63.4 &61.5 &65.2 &62.1 &59.2 &65.1\\
SgMg~\cite{SgMg}  & \pub{ICCV'23} &  {65.7} & {63.9} & {67.4}  & {63.3} & {60.6} & {66.0} \\ 
TempCD~\cite{TempCD}&\pub{ICCV'23} &65.8&63.6 &68.0 & 64.6& 61.6& 67.6\\
SOC~\cite{SOC} & \pub{NIPS'23} &66.0 &64.1& 67.9& 64.2 &61.0 &67.4\\
\rowcolor{cyan!10}\textbf{\ours} (ours)&\pub{CVPR'24}&\textbf{67.1}& \textbf{65.0}& \textbf{69.1}&\textbf{64.9} &\textbf{61.7} &\textbf{68.1}\\
\hline 
\end{tabular}
\vspace{-3mm}
\caption{Comparison on Ref-YouTube-VOS and Ref-DAVIS17.
} \label{tab:ytvos_davis}
\vspace{-6mm}
\end{table}

\begin{table}[t!]
\footnotesize
\centering
\setlength{\tabcolsep}{2pt}
\begin{tabular}{r | c | c c c | c c c }
 \rowcolor[gray]{.9}
\hline
 && \multicolumn{3}{c |}{A2D-Sentences} & \multicolumn{3}{c}{JHMDB-Sentences} \\\rowcolor[gray]{.9}
 Method& Reference& mAP &oIoU & mIoU & mAP & oIoU & mIoU \\
\hline
\hline
\multicolumn{8}{c}{Video-Swin-Tiny}\\
\hline
MTTR~\cite{MTTR} &\pub{CVPR'22}& 46.1 & 72.0 & 64.0 & 39.2 & 70.1 & 69.8  \\
ReferFormer~\cite{referformer} &\pub{CVPR'22}&  52.8 & 77.6 & 69.6 & 42.2 & 71.9 & 71.0  \\ 
HTML~\cite{HTML}&\pub{ICCV'23}&53.4 & 77.6 &69.2 &42.7&-&-   \\ 
SOC~\cite{SOC}&\pub{NIPS'23}&54.8 &78.3& 70.6 &42.7 &72.7 &71.6\\
SgMg~\cite{SgMg} &\pub{ICCV'23}& {56.1} & {78.0} & {70.4} & {44.4} & {72.8} & {71.7} \\ 
\rowcolor{cyan!10}\textbf{\ours} (ours)&\pub{CVPR'24}&\textbf{57.2}&\textbf{79.0}&\textbf{71.3}&\textbf{44.9}&\textbf{73.1}&\textbf{72.1} \\
\hline
\multicolumn{8}{c}{Video-Swin-Base}\\
\hline
ReferFormer~\cite{referformer} &\pub{CVPR'22}&  55.0 & 78.6 & 70.3 & 43.7 & 73.0 & 71.8\\
OnlineRefer~\cite{OnlineRefer} &\pub{ICCV'23}& -&79.6& 70.5&- &73.5 &71.9\\

HTML~\cite{HTML}&\pub{ICCV'23}& 56.7 & 79.5 &71.2 &44.2&-&-\\ 

SOC~\cite{SOC}&\pub{NIPS'23}&57.3  &80.7 & 72.5 & 44.6 &73.6 &72.3\\
SgMg~\cite{SgMg} &\pub{ICCV'23}&  {58.5} & {79.9} & {72.0} & {45.0} & {73.7} & {72.5}  \\ 
\rowcolor{cyan!10}\textbf{\ours} (ours)&\pub{CVPR'24}&\textbf{59.8}&\textbf{81.1}&\textbf{72.9}&\textbf{45.8}&\textbf{73.9}&\textbf{73.0}\\
\hline
\end{tabular}
\vspace{-3mm}
\caption{Comparison on A2D-Sentences and JHMDB-Sentences.} \label{tab:A2D_JHMDB}
\vspace{-3mm}
\end{table}

\subsection{Qualitative Visualization}

As shown in \cref{fig:visualization}, \ours is able to understand both fleeting motions ``\textit{turning around, falling over}'' and long-term motion ``\textit{playing, pushing}'' and segment the target object precisely. In contrast,
LMPM~\cite{MeViS} tends to identify all the objects in the video and fails to comprehend the motion information. The qualitative results further show the effectiveness of our \ours that can capture temporal information effectively across varying timescales.
\section{Conclusion}
We propose a decoupled static and hierarchical motion perception approach to enhance temporal comprehension for referring video segmentation. The static cues and motion cues provided by the given language expressions are employed to image-level referring segmentation and temporal-level motion identification, respectively. Additionally, our hierarchical motion perception module effectively captures temporal information across varying timescales and can well handle both the fleeting and long-term motions in the language and video. Furthermore, the use of contrastive learning enables the model to distinguish the motions of visually similar objects. The proposed approach consistently achieves new state-of-the-art performance across 5 datasets.

{
    \small
    \bibliographystyle{ieeenat_fullname}
    \bibliography{main}
}

\end{document}